\documentclass[11pt,letterpaper,logo,twocolumn]{style}

\usepackage[numbers]{natbib}
\usepackage{graphicx}
\usepackage{booktabs}
\usepackage{amsmath,amsfonts,amssymb}
\usepackage{subcaption}
\usepackage{multirow}
\usepackage{colortbl}
\usepackage{listings}
\usepackage{xparse}
\usepackage{fontawesome5}
\usepackage{threeparttable}

\graphicspath{{./}{fig/}{figures/}{plot/}{pdf/}{table/}}
\usepackage{amsthm}
\usepackage{tcolorbox}
\tcbuselibrary{skins,breakable}
\tcbuselibrary{listingsutf8}
\usepackage{titletoc}
\usepackage{pifont}
\usepackage{mathtools}
\usepackage{bbm}
\usepackage{makecell}
\usepackage{adjustbox}

\newcommand{\modelname}{AR-Omni}
\newcommand{\V}{\mathcal{V}}

\newcommand{\wNTP}{\mathcal{L}_{\text{wNTP}}}
\newcommand{\Lperc}{\mathcal{L}_{\text{perc}}}

\title{AR-Omni: A Unified Autoregressive Model for Any-to-Any Generation}
\runningtitle{AR-Omni: A Unified Autoregressive Model for Any-to-Any Generation}
\PublicDate{2026-1-22}

\author{%
  {\Authfont
    \textbf{Dongjie Cheng}\textsuperscript{1}\equal \quad
    \textbf{Ruifeng Yuan}\textsuperscript{1}\equal \quad
    \textbf{Yongqi Li}\textsuperscript{1}\advisor\quad
    \textbf{Runyang You}\textsuperscript{1} \\
    \Authfont
    \textbf{Wenjie Wang}\textsuperscript{2} \quad
    \textbf{Liqiang Nie}\textsuperscript{3} \quad
    \textbf{Lei Zhang}\textsuperscript{1} \quad
    \textbf{Wenjie Li}\textsuperscript{1}
  }\\
  {\Affilfont
    \textsuperscript{1} The Hong Kong Polytechnic University \\
    \textsuperscript{2} University of Science and Technology of China \quad
    \textsuperscript{3} Harbin Institute of Technology (Shenzhen) \\
    \texttt{\{dong-jie.cheng,ruifeng.yuan\}@connect.polyu.hk, liyongqi0@gmail.com}
  }
}

\begin{document}

\begin{abstract}
Real-world perception and interaction are inherently multimodal, encompassing not only language but also vision and speech, which motivates the development of ``Omni'' MLLMs that support both multimodal inputs and multimodal outputs. While a sequence of omni MLLMs has emerged, most existing systems still rely on additional expert components to achieve multimodal generation, limiting the simplicity of unified training and inference. Autoregressive (AR) modeling, with a single token stream, a single next-token objective, and a single decoder, is an elegant and scalable foundation in the text domain. Motivated by this, we present AR-Omni, a unified any-to-any model in the autoregressive paradigm without any expert decoders. AR-Omni supports autoregressive text and image generation, as well as streaming speech generation, all under a single Transformer decoder. We further address three practical issues in unified AR modeling: modality imbalance via task-aware loss reweighting, visual fidelity via a lightweight token-level perceptual alignment loss for image tokens, and stability–creativity trade-offs via a finite-state decoding mechanism. Empirically, AR-Omni achieves strong quality across three modalities while remaining real-time, achieving a 0.88 real-time factor for speech generation.
\end{abstract}

\newcommand{\TitleLinks}{%
\centering
    \vspace{6pt}
    {\noindent\absfont\fontsize{11}{13}\selectfont
    \faGithub\ Project Page: \url{https://modalitydance.github.io/AR-Omni}\par}%
}


\maketitle

\section{Introduction}
Large Language Models (LLMs) have achieved strong performance in understanding and generating natural language~\citep{achiam2023gpt}. However, their interface is largely limited to text. In contrast, real-world perception and interaction are inherently multimodal, encompassing not only language but also vision and speech. This interaction gap has motivated Multimodal Large Language Models (MLLMs), which extend LLMs to process multimodal capabilities.
Early MLLMs extend LLMs with strong \emph{multimodal perception}, enabling them to interpret multimodal inputs. A natural next step is to further equip MLLMs with \emph{multimodal generation} capabilities, allowing them to respond to users in multiple modalities. This ability that supports both multimodal inputs and multimodal outputs is often called “Omni”~\citep{hurst2024gpt}.

A sequence of omni MLLMs has progressively emerged. SpeechGPT~\citep{zhang2023speechgpt} enables speech-text outputs by discretizing speech and training for cross-modal conversational instruction following. Chameleon~\citep{meta2024chameleon} extends the same ambition to vision, proposing an early-fusion MLLM over a joint discrete vocabulary that supports both image and text generation. Building toward tri-modal any-to-any interaction, AnyGPT~\citep{zhan2024anygpt} and MIO~\citep{wang2024mio} perform autoregressive modeling paired with expert diffusion decoders, thereby supporting flexible tri-modal understanding and generation.

\newcommand{\mio}[2]{%
  \makebox[0.8cm][r]{#1}\,\textbar\,\makebox[0.8cm][l]{#2}%
}

\begin{table*}[t]
\centering\small
\setlength{\tabcolsep}{4pt}
\renewcommand{\arraystretch}{1.12}
\begin{tabular}{l c c c c}
\toprule
Method & Modality I/O & Diffusion-free & Streaming & Real-time \\
\midrule
Kosmos~~\citep{huang2023language}
  & \mio{T,I}{T}
  & -- & -- & -- \\
Flamingo~~\citep{Alayrac2022FlamingoAV}
  & \mio{T,I}{T}
  & -- & -- & -- \\
Chameleon~~\citep{meta2024chameleon}
  & \mio{T,I}{T,I}
  & $\checkmark$ & -- & -- \\
USLM~~\citep{zhang2023speechtokenizer}
  & \mio{T,S}{S}
  & -- & $\times$ & $\checkmark$ \\
AnyGPT~~\citep{zhan2024anygpt}
  & \mio{T,S,I}{T,S,I}
  & $\times$ & $\times$ & $\times$ \\
MiO~~\citep{wang2024mio}
  & \mio{T,S,I}{T,S,I}
  & $\times$ & $\times$ & $\times$ \\
\rowcolor{black!8}
\textbf{AR-Omni (7B)}
  & \mio{\textbf{T,S,I}}{\textbf{T,S,I}}
  & \textbf{$\checkmark$} & \textbf{$\checkmark$} & \textbf{$\checkmark$} \\
\bottomrule
\end{tabular}
\caption{
High-level comparison of multimodal coverage and real-time streaming capability.
\textbf{Diffusion-free} marks methods that do not rely on an external diffusion decoder for image synthesis.
\textbf{Streaming} indicates early emission of playable audio chunks.
\textbf{Real-time} indicates faster-than-real-time synthesis (RTF$<1$) under our setup.
``--'' denotes not applicable. AR-Omni is the only model that achieves both Unified I/O and Real-time Streaming without external diffusion models.
}
\vspace{-1em}
\label{tab:intro-compare-compact}
\end{table*}

Autoregressive (AR) modeling has proven to be an exceptionally elegant and scalable foundation in the text domain: with a single token stream, a single next-token objective, and a single decoder, it can already support powerful text generation. This simplicity is appealing not only conceptually but also practically, which offers the potential for unified training and inference for omni MLLMs. As summarized in Table \ref{tab:intro-compare-compact}, most existing omni MLLMs still rely on additional expert components (e.g., diffusion-style image decoders or non-AR speech generators) to achieve multimodal generation. Motivated by this, we aim to explore whether an omni model can be built with the same AR purity: a single autoregressive model that natively handles text, images, and speech. 

We present \modelname, a unified any-to-any model in the autoregressive paradigm without any expert decoders. \modelname\ tokenizes text, images, and speech into discrete symbols and integrates them into a single joint vocabulary. After training, \modelname\ supports autoregressive text and image generation, as well as streaming speech generation, all under a single 7B-parameter Transformer backbone.

We further address three practical issues in unified AR modeling:
1) \textbf{Modality imbalance.} Unified AR training can be dominated by certain modalities/tasks, leading to skewed learning. We mitigate this via task-aware reweighting on response tails.
2) \textbf{Visual fidelity.} Autoregressively predicting image tokens may sacrifice perceptual quality in the reconstructed visuals. We introduce a lightweight token-level perceptual alignment loss for image tokens.
3) \textbf{Stability-creativity trade-offs.} Different tasks prefer different decoding behaviors, yet a single decoding strategy can be suboptimal across tasks. We employ a finite-state decoding machine that automatically selects greedy decoding for ASR/TTS and sampling decoding for open-ended generation.

Empirically, AR-Omni achieves strong quality across modalities while remaining real-time: AR-Omni attains 146\,ms first-token latency and 0.88 real-time factor for speech generation, together with competitive objective accuracy (6.5 zero-shot TTS WER on VCTK and 9.4 ASR WER on LibriSpeech test-clean). Section~\ref{sec:res} reports detailed quantitative results and qualitative examples demonstrating these any-to-any capabilities.

The key contributions are summarized:
\begin{itemize}
    \item We introduce \modelname, a unified autoregressive model that supports understanding and generation for text, image, and speech without any external task-specific decoders.
    \item We equip \modelname\ with an efficient speech tokenizer, allowing it to decode audio after only a small number of speech tokens and thereby enabling streaming speech.
    \item We address modality imbalance via task-aware reweighting on the response tail and improve visual fidelity with a token-level perceptual alignment loss for image tokens. We also address the trade-off between stability and creativity through a finite-state decoding machine.
\end{itemize}

\begin{figure*}[h]
\centering
\includegraphics[width=\textwidth]{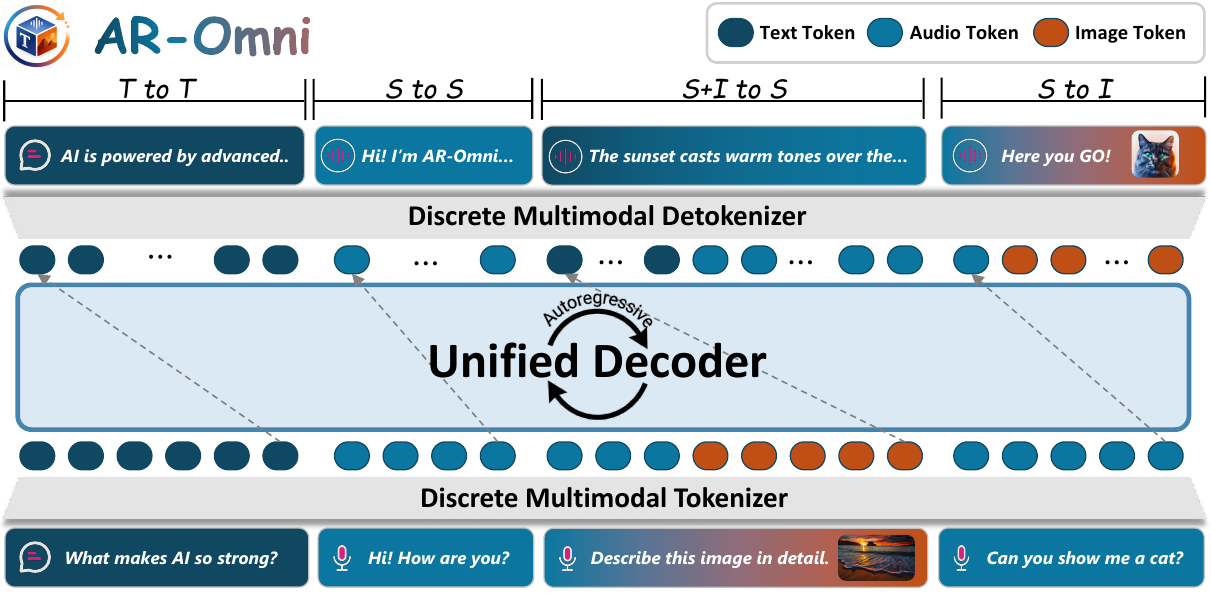}
\caption{Overview of \textbf{AR-Omni}. Text, speech, and image inputs are tokenized and embedded into a shared space. 
A single autoregressive decoder operates over a joint vocabulary to generate a unified token stream. T denotes text, S denotes speech, and I denotes image.}
\vspace{-1em}
\end{figure*}

\section{Related Work}
\subsection{Multimodal Large Language Models}
Early Multimodal Large Language Models (MLLMs)~\citep{li2023blip2,liu2023llava} typically attach modality adapters to Large Language Models (LLMs), enabling multimodal perception and supporting text response. Such approaches have witnessed impressive progress recently, as many more powerful MLLMs~\citep{bai2025qwen2,zhu2025internvl3} have emerged.
Not satisfied with text-only generation, SpeechGPT~\citep{zhang2023speechgpt} discretized speech and enables direct speech understanding and generation. NExT-GPT~\citep{wu2023nextgpt} connected an LLM with modality adaptors and diffusion-based decoders to support multimodal inputs and outputs.
Building toward tri-modal any-to-any interaction, AnyGPT~\citep{zhan2024anygpt} unified text-speech-image understanding and generation. MIO~\citep{wang2024mio} scaled the formulation to video generation under the same paradigm. However, they still rely on an external diffusion decoder for multimodal generation, which means the multimodal modeling responsibility is still mainly taken by external expert models.

\subsection{Discrete Multimodal Tokenization}
The foundation of autoregressive multimodal modeling is to convert continuous modalities into discrete token sequences with vector-quantized (VQ)~\citep{oord2017vqvae} tokenizers. 

\paragraph{Image tokenizers.} Representative designs include VQ-VAE-2~\citep{razavi2019vqvae2}, VQGAN~\citep{esser2021vqgan}, and the dVAE used in DALL$\cdot$E~\citep{ramesh2021dalle}. They compressed images into a 2D grid of discrete codebook indices, which can be flattened for AR training. In practice, the SEED tokenizer~\citep{ge2023seed,ge2023making,rombach2022ldm} produced 1D visual codes and maps these codes to an image embedding that conditions a diffusion UNet to generate pixels.

\paragraph{Speech tokenizers.} Neural audio codecs such as SoundStream~\citep{zeghidour2021soundstream} and EnCodec~\citep{defossez2022encodec} quantized waveforms into discrete codes that can be decoded into high-fidelity audio. 
Another line~\citep{borsos2022audiolm,zhang2023speechtokenizer} factorized speech into semantic tokens for linguistic content and acoustic tokens for timbre. WavTokenizer~\citep{ji2024wavtokenizer} further explored single codebook tokenization for speech to better support streaming speech generation.

\section{\modelname}
\label{sec:method}

\subsection{Unified Autoregressive Modeling}

\modelname\ performs any-to-any generation by mapping all modalities into a shared discrete embedding space, allowing a single transformer-based backbone to generate multimodal data through next-token prediction.
This unification is achieved by the joint vocabulary $\mathcal{V} = \mathcal{V}_{text} \cup \mathcal{V}_{speech} \cup \mathcal{V}_{image}$ that spans all modalities, where each sub-vocabulary contains the discrete tokens for its respective modality. 

\paragraph{Text.}
We utilize the SentencePiece BPE tokenizer from Chameleon~\citep {meta2024chameleon} to tokenize text.

\paragraph{Speech.} 
Unlike prior methods that rely on dual-codebook (semantic and acoustic) tokenizers for easier modeling, we adopt a purely acoustic-based discrete tokenizer~\citep {ji2024wavtokenizer}, eliminating the need for traditional semantic-to-acoustic modeling, thus bypassing the traditional "complete-then-decode" bottleneck and enabling streamed speech responses with low latency.

\paragraph{Image.}
Images are mapped into a sequence of discrete visual codes using a scene-aware VQ tokenizer~\citep{meta2024chameleon}. 
Capturing essential geometric and semantic structures of the visual input in a causal 1D order, this aligns the visual generation process with the language and speech pathways without external diffusion decoders.

\paragraph{Interleaved modeling.}
To integrate these heterogeneous tokens into a single stream, we introduce a set of special tokens to manage modality transitions. Specifically, all tokens are concatenated into a single interleaved sequence $x$, where speech tokens are bracketed by \texttt{<boa>} and \texttt{<eoa>}, and image tokens by \texttt{<boi>} and \texttt{<eoi>}. Additionally, we use \texttt{<eoh>} to indicate the end of the input. For multi-turn dialogues, we use \texttt{<eom>} to mark the end of the model's response for the current turn, and \texttt{<eos>} to mark the end of the entire conversation.
Text acts as the bridge that maintains semantic continuity across these markers. Prompt templates, marker usage, and examples are elaborated in Appendix~\ref{sec:prompt}. This unified formatting reformulates any-to-any generation to a causal probabilistic modeling task:
\begin{equation}
p_\theta(\mathbf{x})=\prod_{t=1}^{T} p_\theta(x_t \mid \mathbf{x}_{<t}),
\end{equation}

At inference time, our decoding strategy is task-aware: we use greedy decoding for deterministic subtasks such as Automatic Speech Recognition (ASR) and Text-to-Speech (TTS), and sampling for open-ended generation such as Text-to-Image generation (T2I). The effectiveness is further discussed in Section~\ref{analy:training}.

\begin{table*}[t]
\centering\small
\setlength{\tabcolsep}{5pt}
\begin{tabular}{p{4.8cm}p{2.2cm}<{\centering}p{2.0cm}<{\centering}p{3.0cm}<{\centering}}
\toprule
Method & Diffusion-free & CIDEr $\uparrow$ & Modality I/O \\
\midrule
Kosmos~~\citep{huang2023language}               & N/S          & 84.7  & T,I \,|\, T \\
Flamingo (9B)~~\citep{Alayrac2022FlamingoAV}    & N/S          & 79.4  & T,I \,|\, T \\
Flamingo (80B)                                  & N/S          & 84.3  & T,I \,|\, T \\
AnyGPT (8B)~~\citep{zhan2024anygpt}             & $\times$     & 107.5 & T,S,I \,|\, T,S,I \\
MIO (7B)~~\citep{wang2024mio}                   & $\times$     & 120.4 & T,S,I \,|\, T,S,I \\
\midrule
Chameleon (7B)~~\citep{meta2024chameleon}       & $\checkmark$ & 13.72 & T,I \,|\, T,I \\
Anole (7B)~~\citep{chern2024anole}              & $\checkmark$ & 15.07 & T,I \,|\, T,I \\
AR-Omni (7B)                                    & $\checkmark$ & 56.53 & T,S,I \,|\, T,S,I \\
\bottomrule
\end{tabular}
\caption{
Image captioning on the MS-COCO Karpathy test split. We report CIDEr.
``Diffusion-free'' indicates whether a model relies on an external diffusion decoder for image generation; N/S means the model does not support image generation.
Modality I/O reports supported input (left) and output (right) modalities, separated by ``|''; \{T,S,I\} denote \{text, speech, image\}.
}

\label{tab:coco}
\end{table*}

\subsection{Optimization}
We employ a composite optimization strategy to address the challenges of multimodal generation. Specifically, we combine the {Weighted Next-Token Prediction} (Weighted NTP) objective to balance modality imbalance with a {Perceptual Loss} (PL) that encourages geometric consistency in the latent  visual space.

\paragraph{Weighted NTP.} 
To mitigate modality imbalance caused by heterogeneous token budgets in tokenized sequences (e.g., speech vs.\ text), we compute the loss as:
\begin{equation}
\wNTP \;=\; -\frac{1}{T}\sum_{t=1}^{T} w_t \,\log p_\theta(x_t \mid x_{<t}),
\end{equation}
where $w_t$ is a scalar weight. In practice, for X2T tasks, including ASR and image captioning, we assign larger weights to the response text tokens. This amplifies supervision on the specific text outputs and reduces the tendency of modalities with longer sequences to dominate optimization. Experimental results in Section~\ref{analy:training} demonstrate the effectiveness of the reweighting strategy.

\paragraph{Perceptual loss.} While standard cross-entropy is effective for exact token matching, it lacks geometric awareness in the discrete latent space: it penalizes all non-target codes equally, ignoring the semantic similarity between visual tokens. To address this, we introduce a perceptual loss that aligns the last-layer hidden states to a frozen, pretrained embedding space of target codes. This guides the model to produce visually coherent structures even when exact token matching fails.

Let $E\in\mathbb{R}^{|V_{image}|\times d_e}$ be the fixed image embedding matrix, $h_t\in\mathbb{R}^{d_h}$ as the last-layer hidden state, and $W_h\in\mathbb{R}^{d_e\times d_h}$ a trainable projection. For timesteps $\mathcal{T}$ whose targets $y_t\in\V_{image}$, we minimize
\begin{equation}
\Lperc \;=\; \frac{1}{|\mathcal{T}|}\sum_{t\in\mathcal{T}}
\big\|\, W_h\,h_t - E[y_t] \,\big\|_2^2,
\end{equation}
so the hidden states are encouraged to match the target code embedding, providing a smoother notion of similarity among visual codes than one-hot supervision.
\paragraph{Total objective.} The total loss is defined as:
\begin{equation}
\mathcal{L} \;=\; \wNTP \;+\; \lambda_{\text{perc}}\,\Lperc,
\end{equation}
with a small $\lambda_{\text{perc}}$ to balance NTP and perceptual gradients.
\paragraph{Training stabilization.} Long interleaved multimodal sequences can be sensitive to optimization.
We use \emph{residual-post-norm} (swin-norm)~\citep{liu2022swin}, which applies normalization on the residual branch.
Let $x\in\mathbb{R}^{T\times d}$ be the input hidden states of a Transformer block. Let $\mathrm{Attn}$ denote multi-head self-attention, $\mathrm{FFN}$ the feed-forward network, and $\mathrm{Norm}$ a normalization operator. The block updates are
\begin{equation}
h \;=\; x + \mathrm{Norm}\,(\mathrm{Attn}(x)).
\end{equation}
\begin{equation}
x' \;=\; h + \mathrm{Norm}\,(\mathrm{FFN}(h)),
\end{equation}
where $h$ is the post-attention state and $x'$ is the block output.
\subsection{Data}
\paragraph{Pre-training data.}\label{sec:pre-training Data} Our pre-training data source is summarized in Table~\ref{tab:pretrain-data}. To maintain a balanced mixture, we sample data with a ratio of $0.5\!:\!1\!:\!2$ for text-only, text-image, and text-speech, respectively. We stop sampling once any modality subset is exhausted, ensuring a controlled modality composition without over-sampling scarce sources. Concretely, we use Ultra-FineWeb~\citep{wang2025ultra} for large-scale text pretraining; LAION and its Aesthetics subset~\citep{schuhmann2022laion} provide broad web-scale image--text pairs for general cross-modal alignment, while JourneyDB~\citep{sun2023journeydb} adds higher-quality image--prompt data to improve text-to-image generation. For speech--text supervision, we use GigaSpeech~\citep{chen2021gigaspeech}, Common Voice~\citep{ardila2020common}, and MLS~\citep{pratap2020mls}. Detailed data description can be found in Appendix~\ref{sec:data_details}.

\begin{table}[t]
\centering
\small
\resizebox{\columnwidth}{!}{%
\begin{tabular}{ll}
\toprule
Modality & Dataset \\
\midrule
Text-only & Ultra-FineWeb~\citep{wang2025ultra} \\
\midrule
Image--Text & LAION-2B~\citep{schuhmann2022laion}, LAION-Aesthetics~\citep{schuhmann2022laion}, 
JourneyDB~\citep{sun2023journeydb} \\
\midrule
Speech--Text & GigaSpeech~\citep{chen2021gigaspeech}, Common Voice~\citep{ardila2020common}, MLS~\citep{pratap2020mls} \\
\bottomrule
\end{tabular}
}
\caption{Summary of pre-training corpora.}
\label{tab:pretrain-data}
\end{table}

\begin{table*}[t]
\centering\small
\begin{tabular}{lcccc}
\toprule
Method     & Decoder Params & Diffusion-free & CLIP$_{score}$  $\uparrow$ & Modality I/O \\
\midrule
GILL~~\citep{Koh2023GeneratingIW}      & 900M  & $\times$ & 0.67 & T,I \,|\, T,I \\
Emu~~\citep{sun2023emu}               & 900M  & $\times$ & 0.66 & T,I \,|\, T,I \\
AnyGPT~~\citep{zhan2024anygpt}         & 900M  & $\times$ & 0.65 & T,S,I \,|\, T,S,I \\
MiO~~\citep{wang2024mio}               & 900M  & $\times$ & 0.64 & T,S,I \,|\, T,S,I \\
\midrule
Chameleon (7B)~~\citep{meta2024chameleon} & 41M & $\checkmark$ & N/S  & T,I \,|\, T,I \\
Anole (7B)~~\citep{chern2024anole}    & 41M  & $\checkmark$ & 0.27 & T,I \,|\, T,I \\
AR-Omni (7B)                           & 41M  & $\checkmark$ & 0.24 & T,S,I \,|\, T,S,I \\
\bottomrule
\end{tabular}
\caption{
Text-to-image generation on MS-COCO. We report CLIP$_{score}$ between generated images and reference captions.
Decoder Params count only the modality-specific image generation module; diffusion-based methods use a latent diffusion UNet ($\sim$900M params), while ours uses a VQGAN detokenizer ($\sim$41M).
``Diffusion-free'' uses $\checkmark$/$\times$ to denote without/with external diffusion, and N/S indicates not supported.
Modality I/O reports supported input (left) and output (right) modalities, separated by ``|''; \{T,S,I\} denote \{text, speech, image\}.
}
\label{tab:t2i}
\end{table*}

\begin{table*}[t]
\centering\small
\resizebox{0.75\textwidth}{!}{%
\begin{tabular}{lcccc}
\toprule
Method & WER $\downarrow$ & Speech-in tok/s $\downarrow$ & Codebook & Modality I/O\\
\midrule
Human-level~\citep{amodei2016deep} & 5.8 & N/S & N/S & S \,|\, T \\
Wav2vec 2.0~~\citep{Baevski2020wav2vec2A} & \textbf{{\color[HTML]{9B9B9B} 2.7}} & N/S & N/S & S \,|\, T \\
Whisper Large V2~~\citep{Radford2022RobustSR} & \textbf{2.7} & N/S & N/S & S \,|\, T \\
AnyGPT~~\citep{zhan2024anygpt} & 8.5 & 50 & Dual & T,S,I \,|\, T,S,I \\
MiO~~\citep{wang2024mio} & 6.3 & 200 & Dual & T,S,I \,|\, T,S,I \\
\midrule
Anole (7B)~~\citep{chern2024anole} & N/S & N/S & N/S & T,I \,|\, T,I \\
AR-Omni (7B) & 9.4 & 40 & Single & T,S,I \,|\, T,S,I \\
\bottomrule
\end{tabular}
}
\caption{
ASR performance on LibriSpeech test-clean. We report WER.
Speech-in tok/s is the number of discrete tokens per second produced by the speech tokenizer for the input audio; N/S indicates not supported.
Codebook uses \emph{Dual} for separate semantic and acoustic codebooks and \emph{Single} for a unified codebook.
Modality I/O reports supported input (left) and output (right) modalities, separated by ``|''; \{T,S,I\} denote \{text, speech, image\}.
}
\label{tab:asr}
\end{table*}

\begin{table*}[t]
  \centering\small
  \resizebox{\textwidth}{!}{%
  \begin{tabular}{lccccccc}
  \toprule
  Method & WER $\downarrow$ & Speech-out tok/s $\downarrow$ & FTL (ms) $\downarrow$ & RTF $\downarrow$ & Real-time ($\text{RTF}<1$) & Codebook & Modality I/O \\
  \midrule
  \textit{Ground Truth} & \textit{1.9} & N/S & N/S & N/S & N/S & N/S & T,S \,|\, S \\
  VALL-E~~\citep{Wang2023NeuralCL} & 7.9 & 600 & N/S & N/S & N/S & Dual & T,S \,|\, S \\
  USLM~~\citep{zhang2023speechtokenizer} & \textbf{6.5} & 50 (+350) & 11611 & 0.89 & $\checkmark$ & Dual & T,S \,|\, S \\
  AnyGPT~~\citep{zhan2024anygpt} & 8.5 & 50 (+350) & 6602 & 2.92 & $\times$ & Dual & T,S,I \,|\, T,S,I \\
  MiO~~\citep{wang2024mio} & 12.0 & 200 & 29079 & 48.90 & $\times$ & Dual & T,S,I \,|\, T,S,I \\
  \midrule
  Anole (7B)~~\citep{chern2024anole} & N/S & N/S & N/S & N/S & N/S & N/S & T,I \,|\, T,I \\
  AR-Omni (7B) & \textbf{6.5} & 40 & \textbf{146} & \textbf{0.88} & \textbf{$\checkmark$} & Single & T,S,I \,|\, T,S,I \\
  \bottomrule
  \end{tabular}
  }
    \caption{
    Zero-shot TTS results on VCTK. We report WER, first-token latency (FTL), and real-time factor (RTF); N/S indicates not supported.
    Speech-out tok/s measures the length of the generated speech token stream per second; for dual-codebook methods, $a(+b)$ denotes semantic tokens at $a$ tok/s plus acoustic tokens at $b$ tok/s.
    Codebook uses \emph{Dual} for separate semantic and acoustic codebooks and \emph{Single} for a unified codebook.
    Modality I/O reports supported input (left) and output (right) modalities, separated by ``|''; \{T,S,I\} denote \{text, speech, image\}.
    }
  \label{tab:tts}
\end{table*}

\paragraph{Omni-interleaved instruction data.} We build our multimodal interleaved instruction data on AnyInstruct~\citep{zhan2024anygpt} by reformatting it under our unified tokenization and task definitions, preserving its interleaved text--image--speech structure. We further apply speech augmentation by mixing in indoor environmental noises from the DEMAND noise dataset~\citep{thiemann_2013_1227121}.
We additionally incorporate VoiceAssistant-400K~\citep{xie2024mini} to enrich speech-centric dialogues. To match the acoustic characteristics of pre-training, we extract the timbre from AR-Omni-Pretrain and synthesize assistant replies using CosyVoice2~\citep{du2024cosyvoice}, reducing the train--test distribution gap in speech. For text-only conversation, we include UltraChat~\citep{ding2023enhancing} as a text-only dialogue source.

\begin{figure*}[t]
  \centering
  \includegraphics[width=\linewidth]{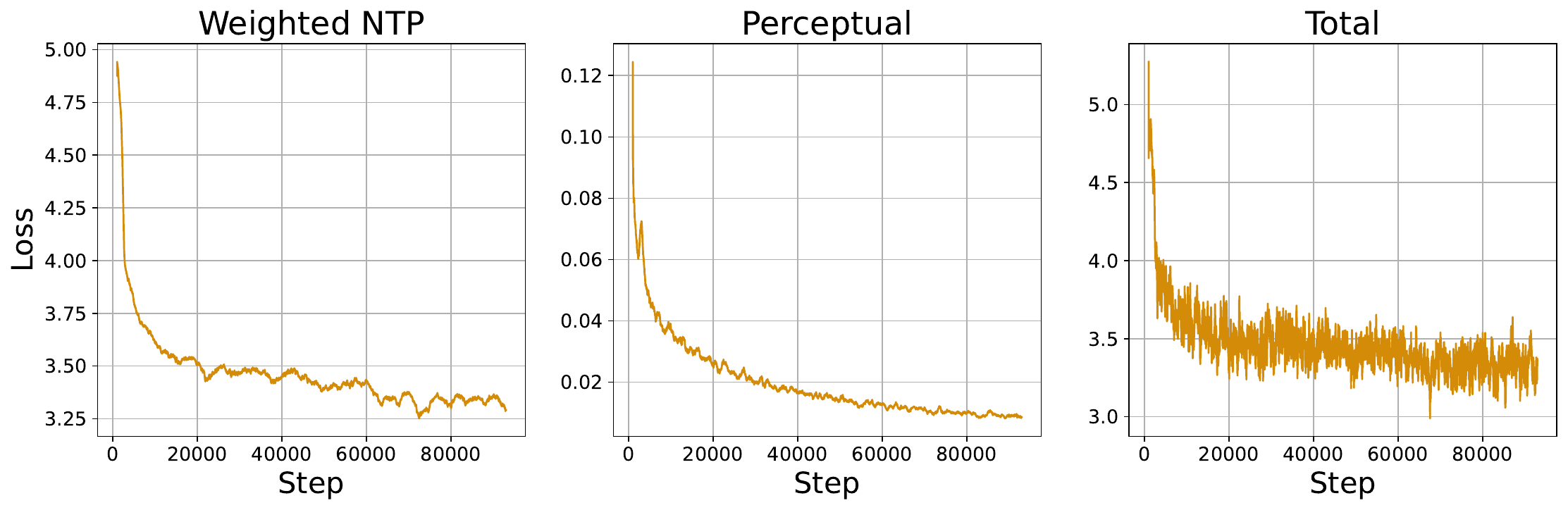}
  \caption{
    Stage~1 pretraining losses of \modelname.
    From left to right: weighted NTP loss, perceptual loss, and total loss.
    Curves are smoothed for readability and plotted over 1k--93k training steps.
  }
  \label{fig:stage1-loss}
\end{figure*}

\section{Experimental Setup}

\subsection{Training}
The training process of AR-Omni consists of two stages: pre-training and fine-tuning. We initialize from Anole~\citep{chern2024anole}, a 7B autoregressive Transformer designed for interleaved image--text modeling. Stage 1 focuses on pre-training by optimizing the weighted NTP objective with the perceptual loss. Stage 2 performs fine-tuning on omni-interleaved instruction data, where the loss is applied exclusively to the response tokens. Further technical details are provided in Appendix~\ref{app:training_details}.

\subsection{Image Evaluation}
\paragraph{Image understanding.}
We evaluated image-to-text understanding on the MS-COCO 2014 captioning benchmark~\citep{Lin2014MicrosoftCC} using the Karpathy test split, following prior work~\citep{Tang2023AnytoAnyGV}.
We reported CIDEr in a zero-shot setting. CIDEr is computed by measuring the cosine similarity between TF-IDF--weighted n-gram vectors of the generated caption and the set of reference captions.

\paragraph{Image generation.}
We evaluated text-to-image generation on MS-COCO by randomly sampling 30k captions from the validation set, following prior protocols~\citep{ge2023making}.
We generated images with AR-Omni and reported CLIP$_{score}$. CLIP$_{score}$ is computed as the cosine similarity between the CLIP image embedding of a generated image and the CLIP text embedding of its caption using CLIP ViT-L, averaged over all samples.

\subsection{Speech Evaluation}
\paragraph{ASR.}
We evaluated Automatic Speech Recognition (ASR) on the LibriSpeech test-clean split~\citep{panayotov2015librispeech}.
We reported Word Error Rate (WER). WER is computed as (substitutions + deletions + insertions) divided by the number of words in the reference transcript, based on Levenshtein alignment.

\paragraph{TTS.}
We evaluated zero-shot text-to-speech (TTS) on the VCTK~\citep{Veaux2017CSTRVC} dataset by conditioning AR-Omni on text prompts. For TTS, WER is computed between the ASR transcript of the synthesized audio and the reference text. We used Whisper-Large-V2~\citep{Radford2022RobustSR} as the transcriber model.
First Token Latency (FTL) is a theoretical lower bound defined as the latency to the first generated speech token that can be immediately decoded into audio. In the dual-codebook scheme, decoding is conditioned on aligned token pairs, so it typically starts once the corresponding tokens from both codebooks are available.
Real Time Factor (RTF) is computed as the total wall-clock synthesis time divided by the duration of the generated audio.

\section{Main Results}
\label{sec:res}
\subsection{Image Results}
\paragraph{Image understanding.}
Table~\ref{tab:coco} presents zero-shot image captioning results.
Within the diffusion-free autoregressive setting, \modelname\ outperforms the Anole~\citep{chern2024anole} initialization on image-to-text generation.
This suggests that extending a diffusion-free AR backbone to an any-to-any training regime does not compromise captioning quality, and can instead strengthen it.

\paragraph{Image generation.}
Table~\ref{tab:t2i} presents text-to-image generation results.
Compared with the Anole initialization, \modelname\ incurs only a slight drop on this benchmark, suggesting that any-to-any training largely preserves the ability to autoregressively generate image tokens.
Meanwhile, diffusion-based systems score higher, consistent with the advantage of employing a diffusion-style image decoder for image synthesis.
These results highlight a clear trade-off: \modelname\ maintains a diffusion-free, single-model generation pipeline, whereas diffusion decoders improve text-to-image quality at the cost of additional expert components.

\subsection{Speech Results}
\paragraph{ASR.}
Table~\ref{tab:asr} presents the ASR results along with the speech tokenization rate for the input audio.
Compared to any-to-any baselines, \modelname\ achieves comparable recognition accuracy while using significantly fewer speech tokens.
This demonstrates that a single-codebook, low-rate speech tokenization can serve as an effective interface for integrating speech into a unified AR model, without requiring separate modality-specific generators.

\paragraph{TTS.}
Table~\ref{tab:tts} reports zero-shot TTS results, including streaming-related metrics.
Under the same evaluation setup, \modelname\ matches the best-performing token-based baseline in intelligibility, while exhibiting more favorable streaming characteristics in latency and throughput. In contrast, prior any-to-any systems often incur higher latency and slower generation, reflecting the overhead of heavier or more complex generation pipelines.

\begin{table}[t]
  \centering
  \small
  \resizebox{\columnwidth}{!}{%
  \begin{tabular}{lcccc}
    \toprule
    Method & I2T $\uparrow$ & T2I $\uparrow$ & ASR $\downarrow$ & TTS $\downarrow$ \\
    \midrule
    AR-Omni & 0.5022 & 0.2406  & 0.1535   & 0.1105 \\
    w/o PL & 0.4874 & 0.2356 & 0.1395   & 0.1175 \\
    w/o swin-norm & 0.5488 & 0.2392 & 0.1329   & 0.1633 \\
    \textit{Simple NTP} & 0.5074 & 0.2248 & 0.2270   & 0.1589 \\
    \bottomrule
  \end{tabular}%
  }
  \caption{
  Ablation results on four omni-pretraining tasks at 40k steps. T denotes text, S denotes speech, and I denotes image.
  I2T reports CIDEr on image captioning. T2I reports CLIP$_{score}$ on text-to-image generation. ASR and TTS report WER.
  $\uparrow$ indicates higher is better and $\downarrow$ indicates lower is better.
  }
  \vspace{-1em}
  \label{tab:ablation-omni}
\end{table}

\begin{figure}[t]
  \centering
  \includegraphics[width=1\linewidth]{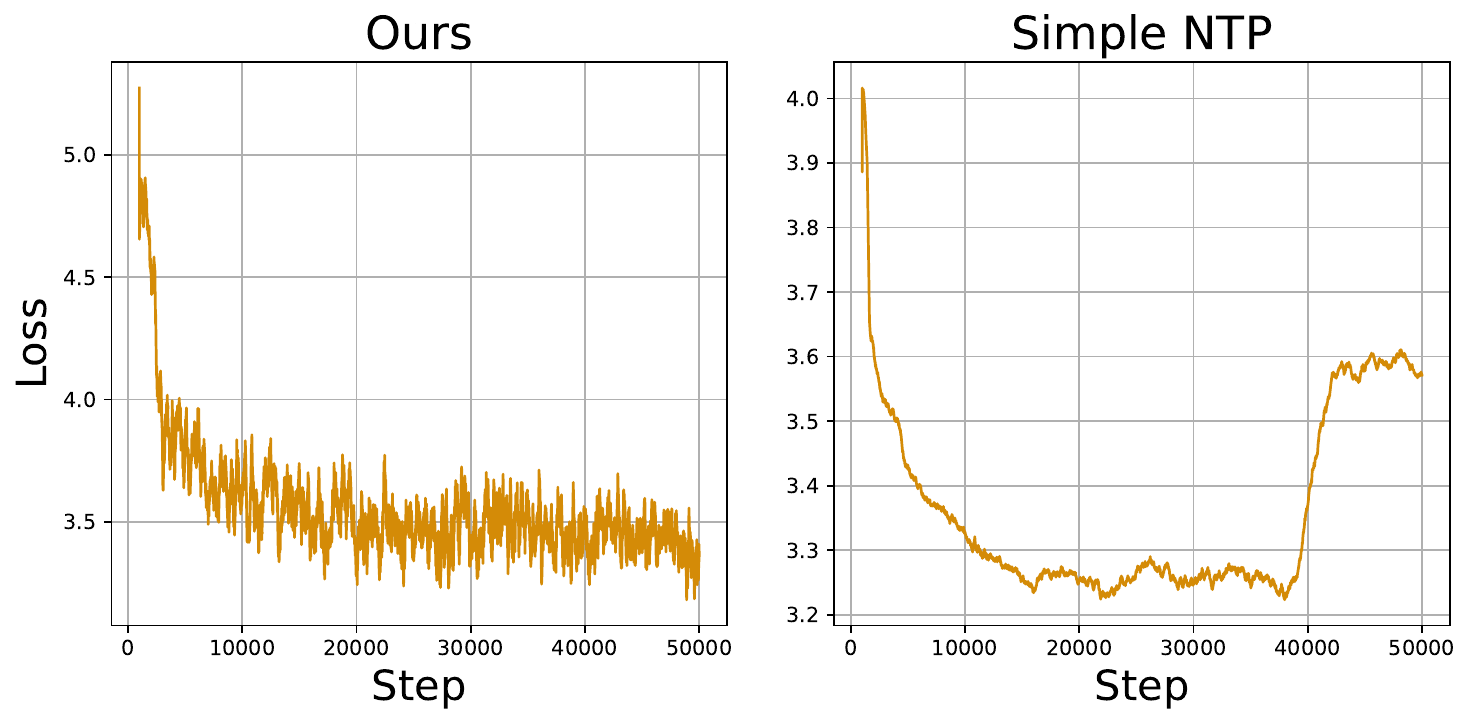}
  \caption{
Loss curves of AR-Omni (\textit{Ours}) and the \textit{simple NTP} training objective.
The naive objective exhibits a sharp loss jump, whereas AR-Omni maintains a smooth and stable loss throughout training.
  }
  \vspace{-1em}
  \label{fig:ablation-loss}
\end{figure}

\begin{figure*}[t]
  \centering
  \includegraphics[width=\textwidth]{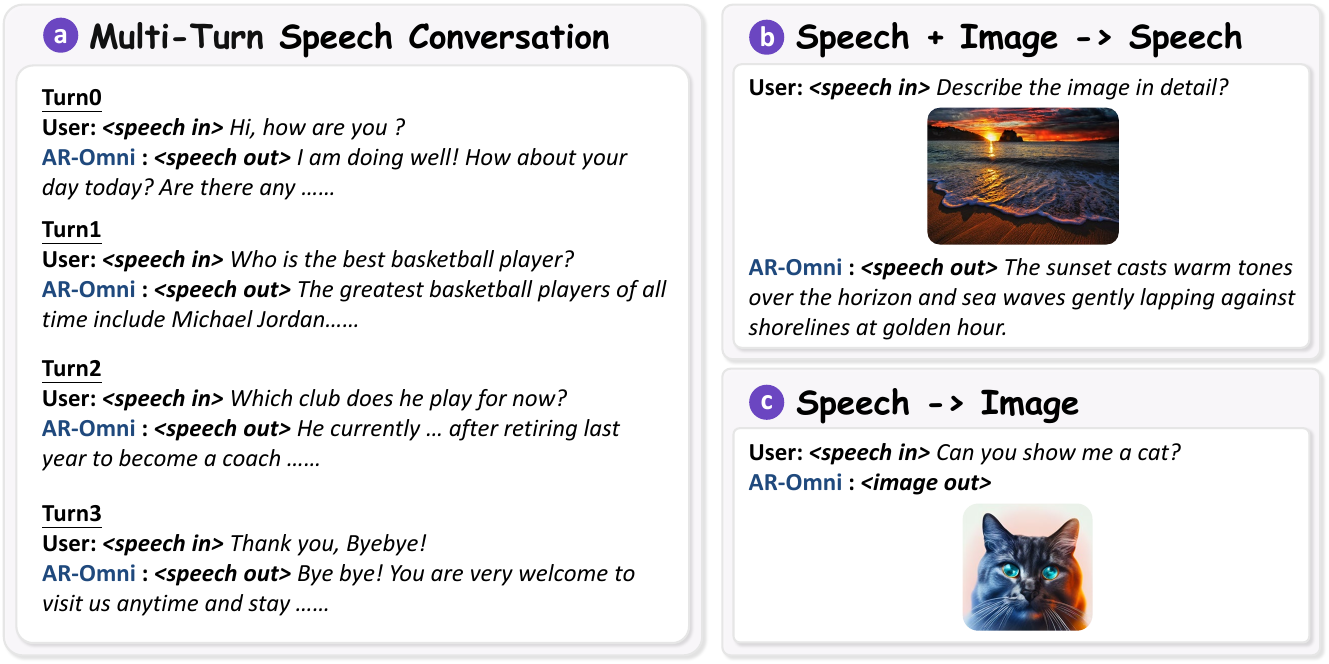}
  \caption{Case studies of AR-Omni: (a) multi-turn speech conversation (S$\rightarrow$S), (b) speech+image understanding with speech response (S+I$\rightarrow$S), and (c) speech-to-image generation (S$\rightarrow$I).}
  \label{fig:case-study}
  \vspace{-0.8em}
\end{figure*}

\section{Further Analysis}

\subsection{Analysis on Training Objectives}
We analyze the stage-1 loss decomposition to understand how the composite pretraining objective is optimized. The results are shown in Figure~\ref{fig:stage1-loss}. The weighted NTP term decreases smoothly throughout training, indicating a stable token-level learning signal under the unified setup. In contrast, the perceptual term converges quickly and saturates at a much smaller magnitude, so the total loss is largely governed by the weighted NTP trajectory; the residual fluctuations are therefore more consistent with mini-batch variance than optimization instability.

\subsection{Ablation Study}
To validate the training design of \modelname\, we conduct ablation analysis focusing on 1) the contribution of individual components to performance, and 2) their impact on long-term training stability.

\paragraph{Component effectiveness.}
We first assess the impact of key components by ablating them at 40k steps, with results summarized in Table~\ref{tab:ablation-omni}.
Removing the perceptual loss slightly reduces I2T and T2I, degrades TTS, and improves ASR. This indicates that PL mainly benefits vision generation and speech synthesis in our unified training framework. Dropping swin-norm improves I2T and ASR but substantially worsens TTS and slightly reduces T2I, suggesting swin-norm is crucial for maintaining speech quality. When removing Weighted NTP, PL, and swin-norm (hereby referred to as \textit{simple NTP}), ASR and TTS errors increase sharply and T2I performance also drops, further demonstrating the effectiveness of the weighted NTP objective for mitigating modality imbalance.

\paragraph{Training stability.} \label{analy:training}
We further investigate the training dynamics of the \textit{Simple NTP} variant over a longer horizon to assess stability. As illustrated in Figure~\ref{fig:ablation-loss}, \textit{Simple NTP} exhibits training instability, characterized by a loss rebound and abrupt spikes in the late stages, indicating a tendency toward model collapse. In contrast, our \modelname\ maintains a smooth and convergent loss trajectory throughout training. These results confirm that our proposed strategies not only enhance multimodal capabilities but also effectively prevent late-stage collapse, ensuring robust training stability.

\subsection{Case Study}
After omni supervised instruction tuning, AR-Omni exhibits omni multimodal capabilities across text, images, and speech, including understanding and generation. Figure~\ref{fig:case-study} presents representative case studies. In (a), AR-Omni maintains a multi-turn speech context and answers follow-up questions speech coherently.
In (b), given a spoken instruction and an image as context, AR-Omni produces a grounded image description in speech; in (c), AR-Omni generates an image from a spoken prompt.
Together, these cases illustrate unified perception, understanding, and generation across heterogeneous modality contexts. Longer-horizon and more fully interleaved demonstrations are provided in Appendix~\ref{sec:case_appendix}.

\section{Conclusion and Future Work}
We presented \modelname, a unified any-to-any model in the autoregressive paradigm without any expert decoders, which tokenizes text, images, and speech into a discrete token stream. With one Transformer backbone, \modelname\ supports autoregressive text and image generation as well as streaming speech generation. To make unified AR modeling practical, we mitigate modality imbalance with task-aware loss reweighting, improve visual fidelity with perceptual loss, and adapt decoding behavior via a finite-state decoding machine. Experiments show competitive tri-modal capabilities while remaining real-time for streaming speech. A key limitation is that diffusion-free autoregressive image generation still lags behind diffusion-based systems in terms of image generation. Future work will focus on enhancing the quality of diffusion-free image generation while respecting the purity of the unified AR.


\bibliographystyle{unsrtnat} 
\bibliography{ref}


\appendix
\clearpage
\section{Training Details}
\label{app:training_details}
We train AR-Omni on 8 NVIDIA A100 GPUs. The pipeline consists of a pre-training stage followed by fine-tuning, both optimized with Adam and a linear learning-rate schedule with warmup. We apply global gradient clipping to stabilize unified multimodal training. Our implementation is developed on top of the Chameleon repository~\citep {meta2024chameleon}, including training and inference code. Table~\ref{tab:training-settings} lists the key hyperparameters for both stages.

\begin{table}[h]
  \centering
  \small
  \renewcommand{\arraystretch}{1.15}
  \begin{tabular}{lcc}
    \toprule
    \textbf{Hyperparameter} & \textbf{Pre-training} & \textbf{Fine-tuning} \\
    \midrule
    GPU Type            & A100        & A100        \\
    Optimizer           & Adam        & Adam        \\
    LR Scheduler        & Linear      & Linear      \\
    Gradient Clipping   & 1.0         & 1.0         \\
    Max Sequence Length & 1300        & 3456        \\
    Warmup Ratio        & 0.05        & 0.05        \\
    Batch Size          & 480         & 64          \\
    Training Steps      & 140{,}000   & 18{,}000    \\
    Peak Learning Rate  & $6\times10^{-5}$ & $2\times10^{-5}$ \\
    \bottomrule
  \end{tabular}
  \caption{Training hyperparameters for pre-training and fine-tuning. Batch size denotes the global batch size.}
  \label{tab:training-settings}
\end{table}

\section{Dataset Details}
\label{sec:data_details}
Table~\ref{tab:pretrain-data-detailed} summarizes the pre-training corpora used for AR-Omni, covering image--text, speech--text, and text-only data. For each dataset, we report modality pairing, scale statistics, and brief notes on source and preprocessing to facilitate reproducibility. We use publicly available datasets and follow standard filtering and deduplication practices to reduce noise and unsafe content.

\begin{table*}[t]
\centering
\footnotesize
\setlength{\tabcolsep}{5pt}
\renewcommand{\arraystretch}{1.15}

\begin{minipage}{\textwidth}
\centering
\begin{tabular}{p{0.23\textwidth} p{0.70\textwidth}}
\toprule
Dataset & Details \\
\midrule

\multicolumn{2}{l}{\textbf{Image--Text}}\\
\midrule
LAION-2B-en &
\textbf{Size:} 2.32 billion image--text pairs\newline
\textbf{Language:} English\newline
\textbf{Content and source:} Web image URLs and associated alternative text collected from Common Crawl\newline
\textbf{Notes:} A Contrastive Language--Image Pre-training filtered subset of LAION-5B; LAION distributes index files (Uniform Resource Locators and metadata) rather than the image files themselves. \\
\midrule

LAION-Aesthetics V2 &
\textbf{Size:} 600 million image--text pairs\newline
\textbf{Language:} English\newline
\textbf{Content and source:} English portion of LAION-5B ranked by a predicted aesthetic score; we select examples with predicted score at least 5\newline
\textbf{Notes:} Aesthetic-score subsets are nested and overlapping. \\
\midrule

JourneyDB &
\textbf{Size:} approximately 4.4 million image--prompt pairs\newline
\textbf{Language:} English prompts\newline
\textbf{Content and source:} High-resolution images generated by Midjourney with their corresponding text prompts\newline
\textbf{Notes:} The dataset also provides additional annotations (for example, captions and visual question answering style annotations) used for evaluation. \\
\midrule

\multicolumn{2}{l}{\textbf{Speech--Text}}\\
\midrule
GigaSpeech &
\textbf{Size:} 10{,}000 hours of transcribed audio \newline
\textbf{Language:} English\newline
\textbf{Content and source:} A mixture of read and spontaneous speech from audiobooks, podcasts, and online videos\newline
\textbf{Notes:} The corpus provides multiple predefined subsets that support training at different scales. \\
\midrule

Common Voice &
\textbf{Size:} 33{,}150 hours total audio, including 22{,}108 hours validated\newline
\textbf{Language:} Multiple languages\newline
\textbf{Content and source:} Crowdsourced speech recordings paired with transcripts; optional speaker metadata is available\newline
\textbf{Notes:} Reported sizes depend on the release version; we report statistics for version 20.0. \\
\midrule

Multilingual LibriSpeech &
\textbf{Size:} 44{,}500 hours of transcribed audio\newline
\textbf{Language:} English\newline
\textbf{Content and source:} Audiobook recordings from LibriVox aligned to transcripts\newline
\textbf{Notes:} We use only the English portion of Multilingual LibriSpeech. \\
\midrule
\multicolumn{2}{l}{\textbf{Text-only}}\\
\midrule
Ultra-FineWeb&
\textbf{Size:} A subset sampled from the English split of Ultra-FineWeb. For reference, the full English split contains approximately one trillion tokens and about 1.16 billion rows.\newline
\textbf{Language:} English\newline
\textbf{Content and source:} High-quality web text derived from Common Crawl through an efficient filtering and verification pipeline applied to FineWeb.\newline
\textbf{Notes:} We use only a portion of the English split and do not use the Chinese split. \\

\bottomrule
\end{tabular}
\end{minipage}

\caption{Detailed statistics of pre-training corpora.}
\label{tab:pretrain-data-detailed}
\end{table*}

\section{Prompt Templates}
\label{sec:prompt}
We adopt a unified dialogue-style prompt format for any-to-any multimodal tasks (Table~\ref{tab:prompt_task_templates}). Each user turn begins with \texttt{<bos>} and ends with \texttt{<eoh>}. Non-text modalities are explicitly bracketed by boundary tokens, including \texttt{<boa>}/\texttt{<eoa>} for audio tokens and \texttt{<boi>}/\texttt{<eoi>} for image tokens. For single-turn (non-chat) tasks, the assistant response is terminated by \texttt{<eos>}. For multi-turn chat settings, the assistant response is terminated by \texttt{<eom>}, and the dialogue history is formed by concatenating previous turns in the same format.

This unified formatting casts diverse tasks into a single next-token prediction interface while keeping modality boundaries explicit for both inputs and outputs.

\begin{table*}[!htbp]
    \centering
    \setlength{\tabcolsep}{10pt}
    \renewcommand{\arraystretch}{1.2}
    \newcommand{\task}[1]{\multicolumn{1}{c}{#1}\\}
    \begin{tabularx}{\textwidth}{X}
        \toprule
        \textbf{Prompt Templates:}\\[2pt]
        \midrule

        \task{Image Generation (Text $\rightarrow$ Image)}
        \midrule
        \texttt{<bos>} \textbf{USER}: Create an image for: \textbf{\{caption\}} \texttt{<eoh>} \\
        \textbf{ASSISTANT}: \texttt{<boi>} \textbf{\{image\}} \texttt{<eoi>} \texttt{<eos>}\\[4pt]
        \midrule

        \task{ASR (Audio $\rightarrow$ Text)}
        \midrule
        \texttt{<bos>} \textbf{USER}: Transcribe the audio. \texttt{<boa>} \textbf{\{audio\}} \texttt{<eoa>} \texttt{<eoh>} \\
        \textbf{ASSISTANT}: \textbf{\{transcript\}} \texttt{<eos>}\\[4pt]
        \midrule

        \task{TTS (Text $\rightarrow$ Audio)}
        \midrule
        \texttt{<bos>} \textbf{USER}: Convert the text to speech: \textbf{\{text\}} \texttt{<eoh>} \\
        \textbf{ASSISTANT}: \texttt{<boa>} \textbf{\{audio\}} \texttt{<eoa>} \texttt{<eos>}\\[4pt]
        \midrule

        \task{Image Captioning (Image $\rightarrow$ Text)}
        \midrule
        \texttt{<bos>} \textbf{USER}: Describe the image. \texttt{<boi>} \textbf{\{image\}} \texttt{<eoi>} \texttt{<eoh>} \\
        \textbf{ASSISTANT}: \textbf{\{caption\}} \texttt{<eos>}\\[4pt]
        \midrule

        \task{Chat-Image Generation (Audio $\rightarrow$ Text $\rightarrow$ Image)}
        \midrule
        \textbf{\{history\}} \\
        \texttt{<bos>} \textbf{USER}: \texttt{<boa>} \textbf{\{audio\}} \texttt{<eoa>} Transcribe the audio and generate an image for the transcript. \texttt{<eoh>} \\
        \textbf{ASSISTANT}: \textbf{\{transcript\}} \texttt{<boi>} \textbf{\{image\}} \texttt{<eoi>} \texttt{<eom>}\\[4pt]
        \midrule

        \task{Chat-Speech Conversation (Audio $\rightarrow$ Text)}
        \midrule
        \textbf{\{history\}} \\
        \texttt{<bos>} \textbf{USER}: \texttt{<boa>} \textbf{\{audio\}} \texttt{<eoa>} Reply in text. \texttt{<eoh>} \\
        \textbf{ASSISTANT}: \textbf{\{response\}} \texttt{<eom>}\\
        \bottomrule
    \end{tabularx}
    \caption{Dialogue-style prompt templates with explicit modality boundaries. Single-turn tasks terminate assistant outputs with \texttt{<eos>}, while chat settings terminate each assistant message with \texttt{<eom>}. \textbf{\{history\}} denotes a concatenation of previous turns in the same format.}
    \label{tab:prompt_task_templates}
\end{table*}

\section{Case Study}
\label{sec:case_appendix}
\begin{figure*}[t]
  \centering
  \includegraphics[width=0.9\textwidth]{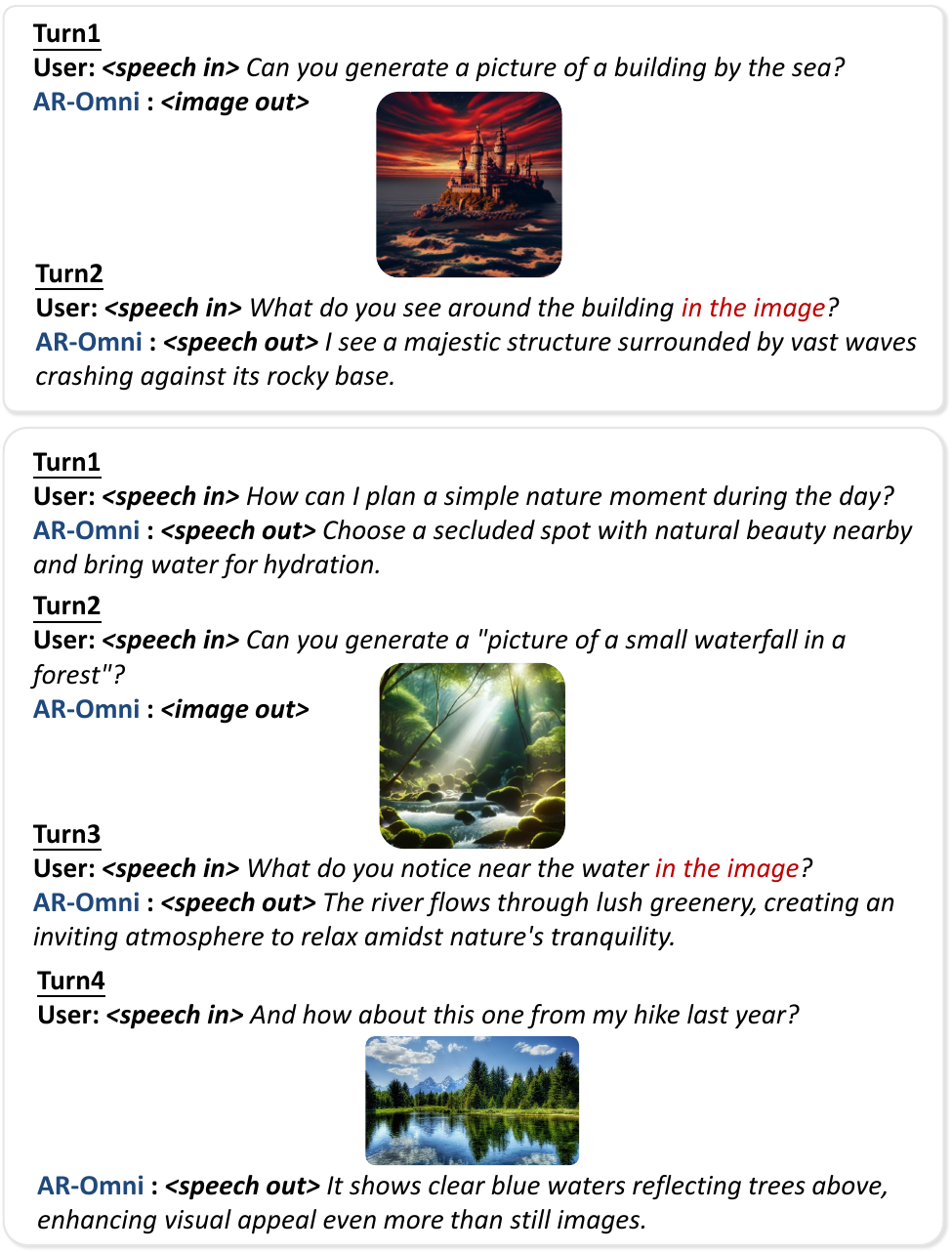}
  \caption{Multi-turn interleaved conversation example.}
  \label{fig:appendix-interleaved}
\end{figure*}

\begin{figure*}[t]
  \centering
  \includegraphics[width=0.9\textwidth]{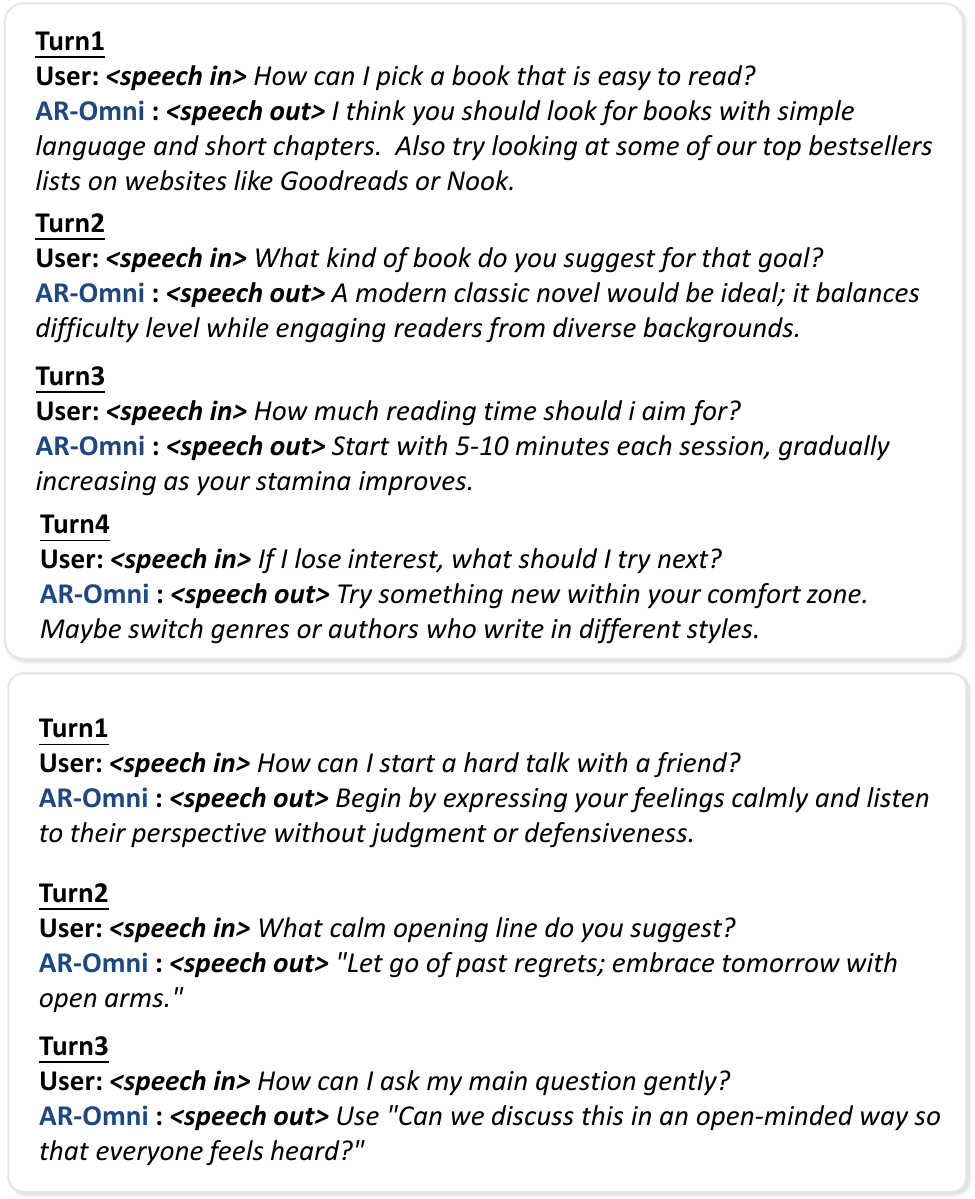}
  \caption{Multi-turn speech conversation example.}
  \label{fig:appendix-speech}
\end{figure*}

\begin{figure*}[t]
  \centering
  \includegraphics[width=0.8\textwidth]{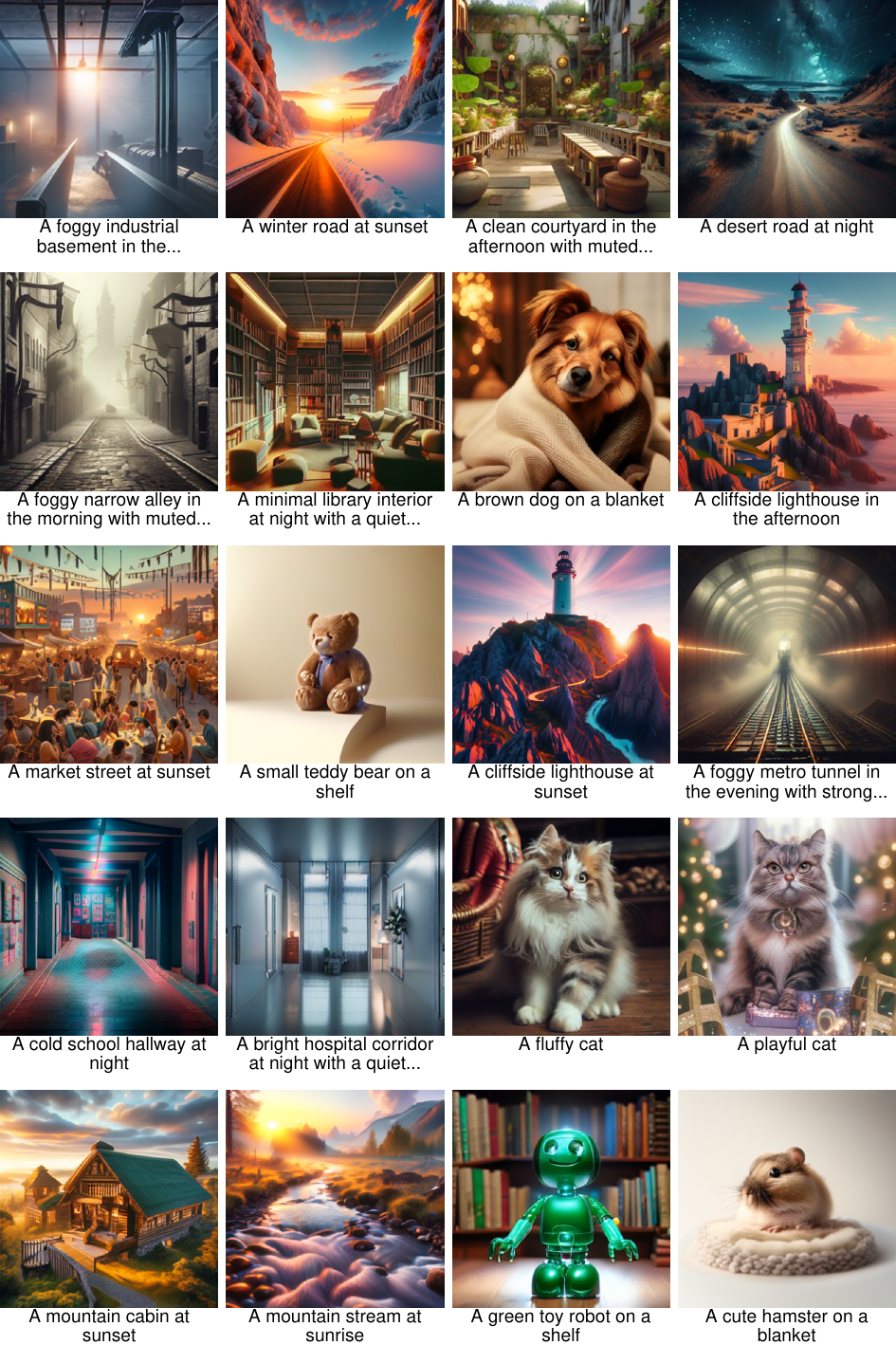}
  \caption{Qualitative image generation results with \modelname\ across diverse prompts and styles.}
  \label{fig:mosaic-001}
\end{figure*}

\begin{figure*}[t]
  \centering
  \includegraphics[width=0.8\textwidth]{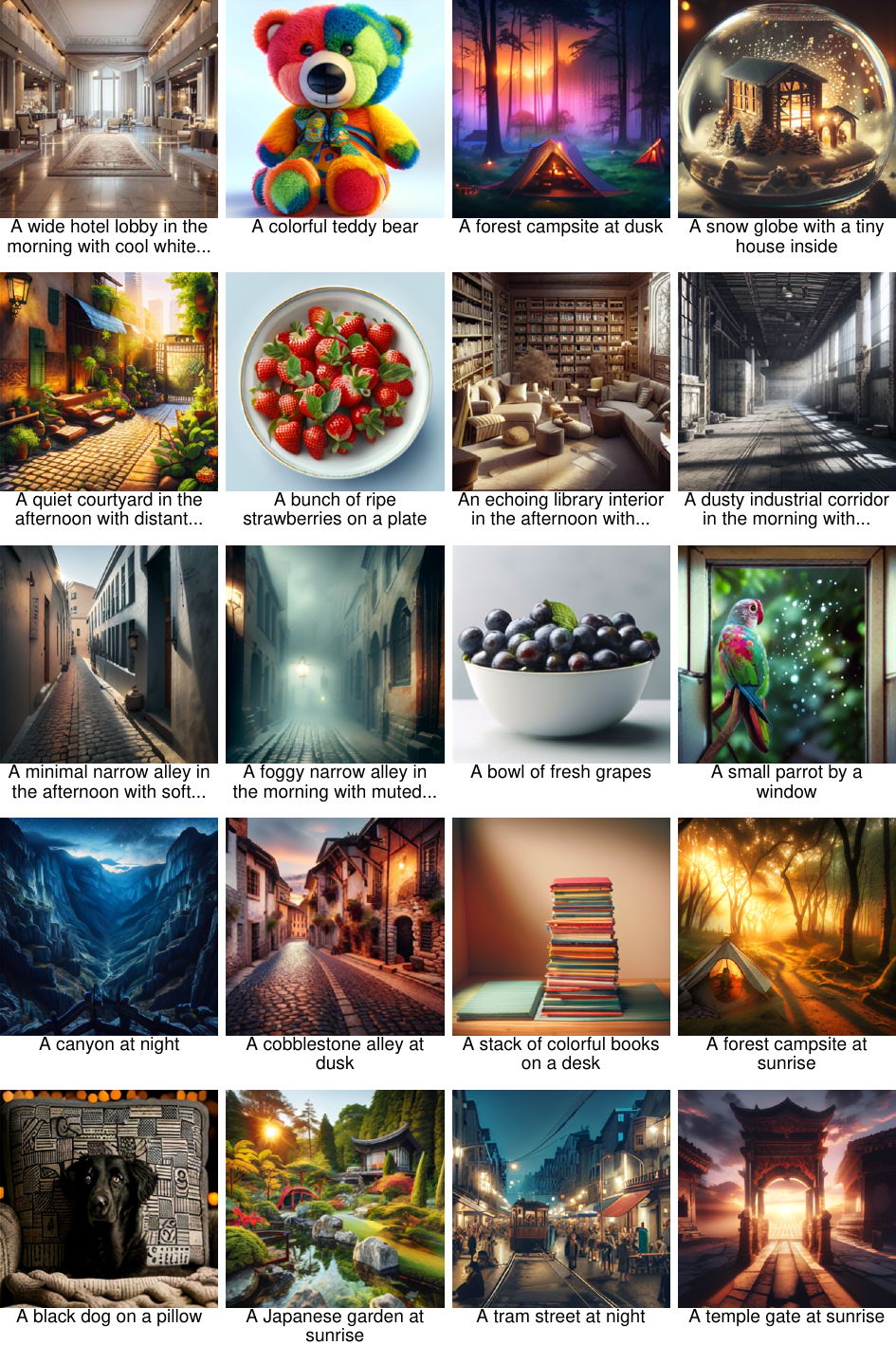}
  \caption{Qualitative image generation results with \modelname\ across diverse prompts and styles.}
  \label{fig:mosaic-002}
\end{figure*}

\begin{figure*}[t]
  \centering
  \includegraphics[width=0.8\textwidth]{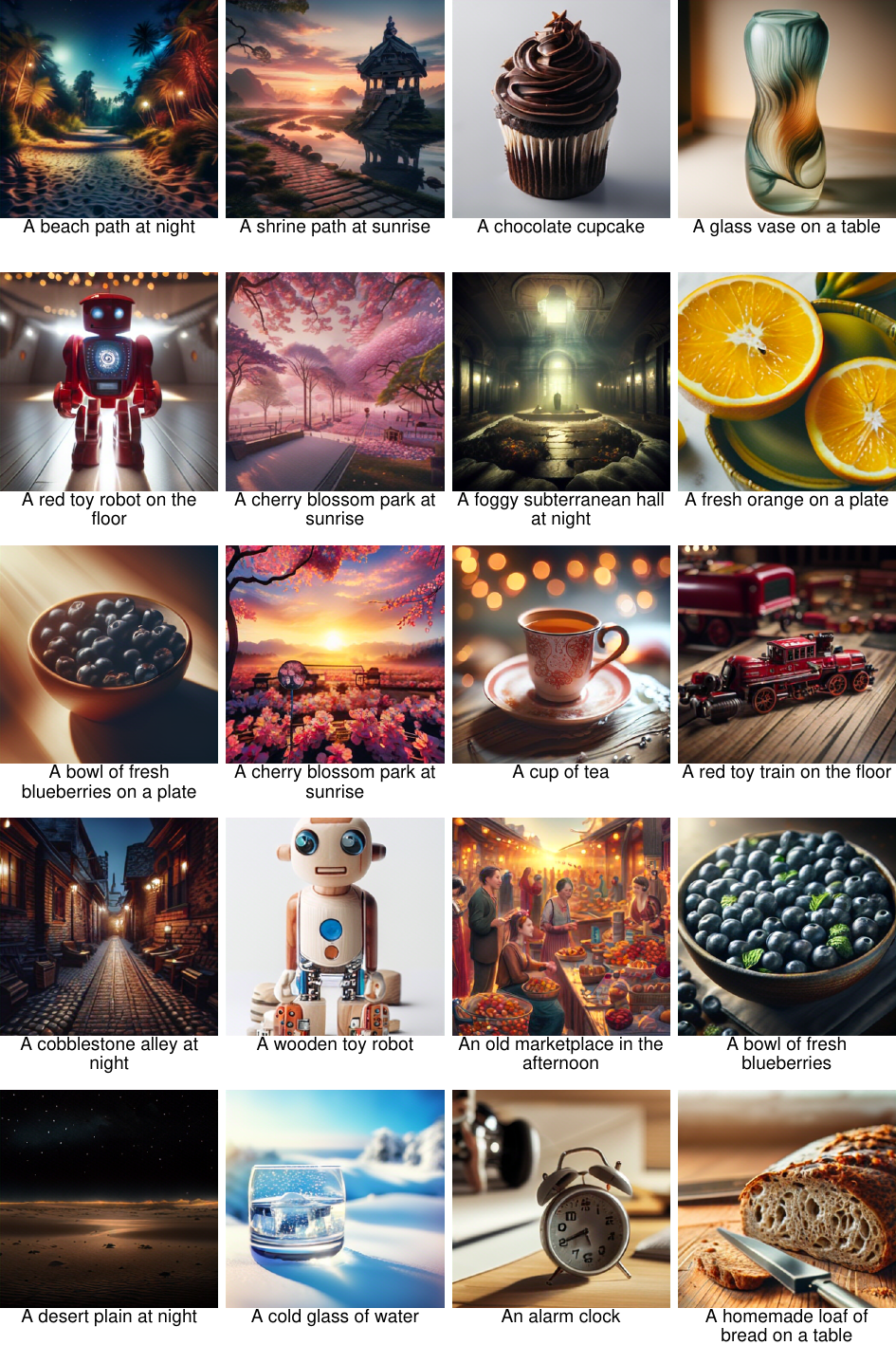}
  \caption{Qualitative image generation results with \modelname\ across diverse prompts and styles.}
  \label{fig:mosaic-003}
\end{figure*}

\begin{figure*}[t]
  \centering
  \includegraphics[width=0.8\textwidth]{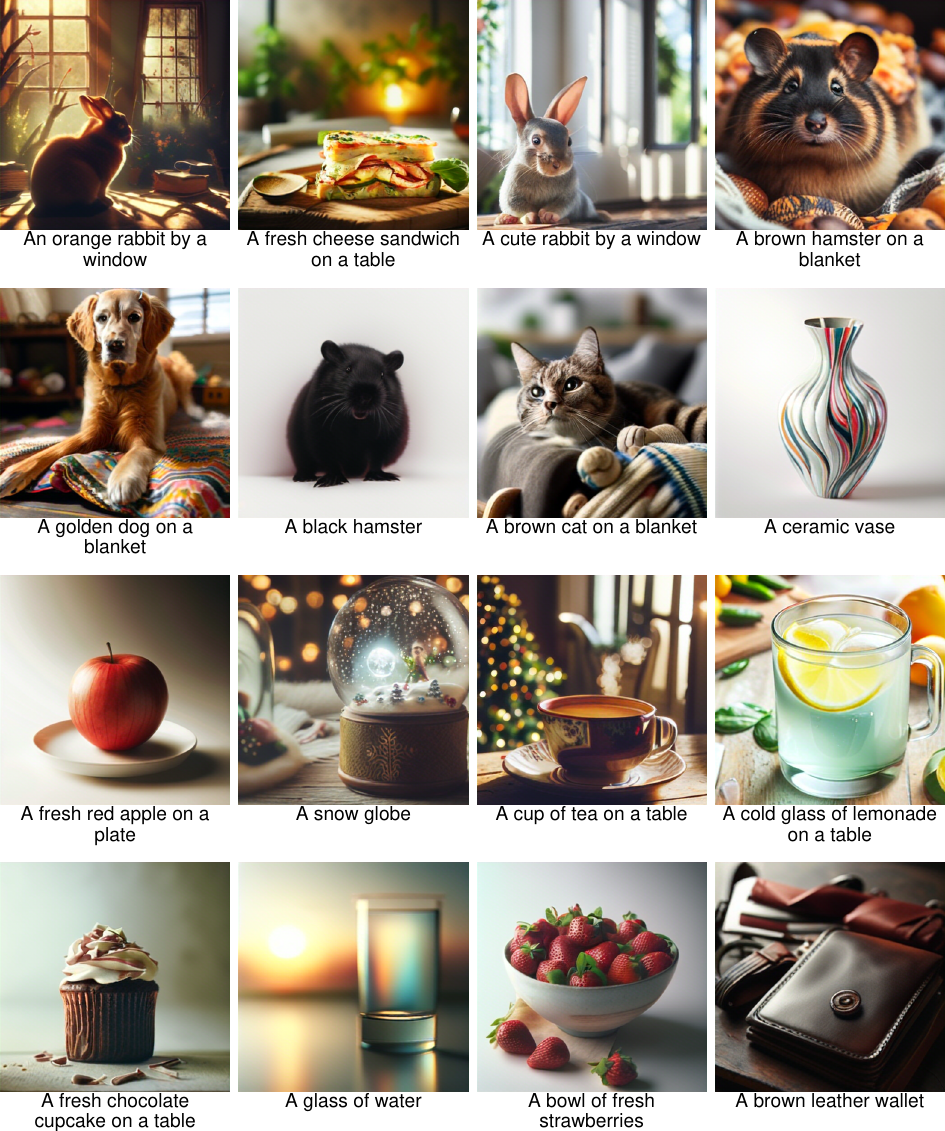}
  \caption{Qualitative image generation results with \modelname\ across diverse prompts and styles.}
  \label{fig:mosaic-004}
\end{figure*}

We provide additional qualitative examples illustrating AR-Omni's any-to-any behavior. Figures~\ref{fig:appendix-interleaved} and~\ref{fig:appendix-speech} show multi-turn interactions with interleaved multimodal inputs and outputs. Figures~\ref{fig:mosaic-001}--\ref{fig:mosaic-004} present text-to-image samples generated by AR-Omni without external diffusion decoders; images are obtained by decoding the autoregressively generated discrete image tokens with the detokenizer.

\end{document}